\documentclass[sigconf]{acmart}

\usepackage{xcolor}
\usepackage{pifont}
\usepackage{mathrsfs}
\usepackage{url}
\usepackage{seqsplit}
\usepackage{amsmath}
\newcommand{\cmark}{\textcolor{green}{\ding{51}}}
\newcommand{\xmark}{\textcolor{red}{\ding{55}}}

\AtBeginDocument{%
  }

\copyrightyear{2026}
\acmYear{2026}
\setcopyright{cc}
\setcctype{by}
\acmConference[MM '26]{Proceedings of the 34th ACM International Conference on Multimedia}{November 10--14, 2026}{Rio de Janeiro, Brazil.}
\acmBooktitle{Proceedings of the 34th ACM International Conference on Multimedia (MM '26), November 10--14, 2026, Rio de Janeiro, Brazil}
\acmISBN{979-8-4007-2213-4/2026/11}
\acmDOI{10.1145/3767308.3835378}
\begin{document}

\title{I Seek You in Videos: Identity-Conditioned Queries for Person-Centric Video Reasoning}


\author{Shibo Gao}
\authornote{Both authors contributed equally to this research.}
\affiliation{%
  \institution{Beijing Jiaotong University}
  \city{Beijing}
  \country{China}}
\affiliation{%
  \institution{HUJING Digital Media \& Entertainment Group}
  \city{Beijing}
  \country{China}}
\affiliation{%
  \department{MAIS}
  \institution{Institute of Automation, Chinese Academy of Sciences}
  \city{Beijing}
  \country{China}}

\author{Chongxiao Wang}
\authornotemark[1]
\author{Chenglong Huang}
\author{Jie Ma}
\author{Haolin Shi}
\author{Fei Ding}
\author{Jing Li}
\affiliation{%
  \institution{HUJING Digital Media \& Entertainment Group}
  \city{Beijing}
  \country{China}}

\author{Qiang Lyu}
\affiliation{%
  \department{School of Computer Science and Technology}
  \institution{University of Chinese Academy of Sciences}
  \city{Beijing}
  \country{China}}

\author{Yangyang Liu}
\affiliation{%
  \department{MAIS}
  \institution{Institute of Automation, Chinese Academy of Sciences}
  \city{Beijing}
  \country{China}}

\author{Yang Liu}
\affiliation{%
  \department{College of Electronic and Information Engineering}
  \institution{Tongji University}
  \city{Shanghai}
  \country{China}}

\author{Jun Liu}
\affiliation{%
  \department{School of Computing and Communications}
  \institution{Lancaster University}
  \city{Lancaster}
  \country{United Kingdom}}

\author{Linlin Huang}
\affiliation{%
  \institution{Beijing Jiaotong University}
  \city{Beijing}
  \country{China}}

\author{Peipei Yang}
\correspondingauthor
\affiliation{%
  \department{MAIS}
  \institution{Institute of Automation, Chinese Academy of Sciences}
  \city{Beijing}
  \country{China}}

\renewcommand{\shortauthors}{Shibo Gao et al.}

\begin{abstract}
Real-world video reasoning often involves multimodal, multi-source inputs, whereas existing video reasoning tasks typically assume a simplified video-text setting, limiting identity matching and person-centric reasoning.
To bridge this gap, we introduce the Identity-conditioned Queries (ICQ) task, in which models are required to jointly associate and interpret an input video and a reference image of a person, and leverage this conditioning to address identity grounding, behavior understanding, and temporal reasoning, among other challenges.
Building on ICQ, we present ISYV (I Seek You in Videos), a systematic solution comprising three components: (1) ISYV-Bench, a challenging evaluation benchmark with 1,377 real-world complex videos and 1,377 question–answer pairs, organized into six difficulty levels spanning capabilities from identity recognition to causal reasoning; (2) ISYV-75K, a large-scale training set of 75K high-quality samples constructed via automated annotation, multi-stage verification, and manual review; and (3) ISYV-Framework, containing an ICQ-oriented model and training strategy for learning to exploit informative video shots without additional shot-level annotations.
Extensive experiments show that both mainstream closed-source and open-source MLLMs struggle on ISYV-Bench, especially in cross-domain identity matching and long-horizon tracking. ISYV-Model outperforms strong baselines and in some aspects approaches closed-source performance. Overall, ISYV provides a unified task definition, scalable datasets/benchmarks, and modeling insights for person-centric video reasoning.
\end{abstract}

\ccsdesc[500]{Computing methodologies~Computer vision tasks}

\keywords{MLLM, Video Reasoning, Post-training, Benchmark, Framework}

\maketitle

\section{Introduction}

With the rapid progress of Multimodal Large Language Models (MLLMs) in recent years, video reasoning has become a key task for evaluating models’ higher-order cognitive capabilities and has emerged as an active research area~\cite{qwen25vl,qwen3vl,crossvid,gemini25,videollama3}. Existing work on video reasoning spans major sub-tasks such as video captioning, video question answering, and video grounding~\cite{b-mvbench,b-tvqa,b-vagu,b-activitynetqa,b-nextqa}. However, most studies still follow the traditional “video–text” bimodal paradigm, where models only process video content and generate textual outputs~\cite{kimi,glm41v,internvl3}. The conventional setting overlooks the heterogeneous, multi-source inputs commonly encountered in real applications. For example, users often issue queries conditioned not only on the video itself but also on reference images and specific object descriptions. This gap between simplified formulations and practical demands limits the applicability of current models in complex real-world scenarios.

Identity-conditioned Queries (ICQ) is a challenging person-centric video reasoning task. Unlike conventional video reasoning~\cite{b-mmvu,b-longvideobench}, ICQ requires models to reason over both a complex video and a reference image of a target person. Since the reference image may come from a different source, large appearance gaps in clothing, lighting, makeup, and other factors make identity matching difficult. Models must therefore identify the same person across these variations while handling tasks such as localization, action recognition, temporal grounding, and causal reasoning. Overall, ICQ presents three key challenges: 
\textbf{(i)} integrating heterogeneous inputs, 
\textbf{(ii)} localizing and tracking the target person in complex scenes, 
\textbf{(iii)} cross-domain identity matching with identity-aware understanding. These abilities are important for real-world applications such as video retrieval, personalized analytics, film production, and intelligent surveillance. Although some prior works explore related directions~\cite{ida-vlm,plvm,myvlm,iir-vlm}, they focus mainly on image-level interactions and do not address video-image-text reasoning or provide high-quality benchmarks.

To this end, we propose ISYV (I Seek You in Videos), a systematic solution for the ICQ task, comprising three components: ISYV-Bench, an ICQ-specific evaluation benchmark; ISYV-75K, a large-scale training set; and ISYV-Framework, a dedicated model with an ICQ-oriented training strategy, detailed through our contributions below.

\begin{figure}[t]
	\centering
	\includegraphics[width=0.9\linewidth]{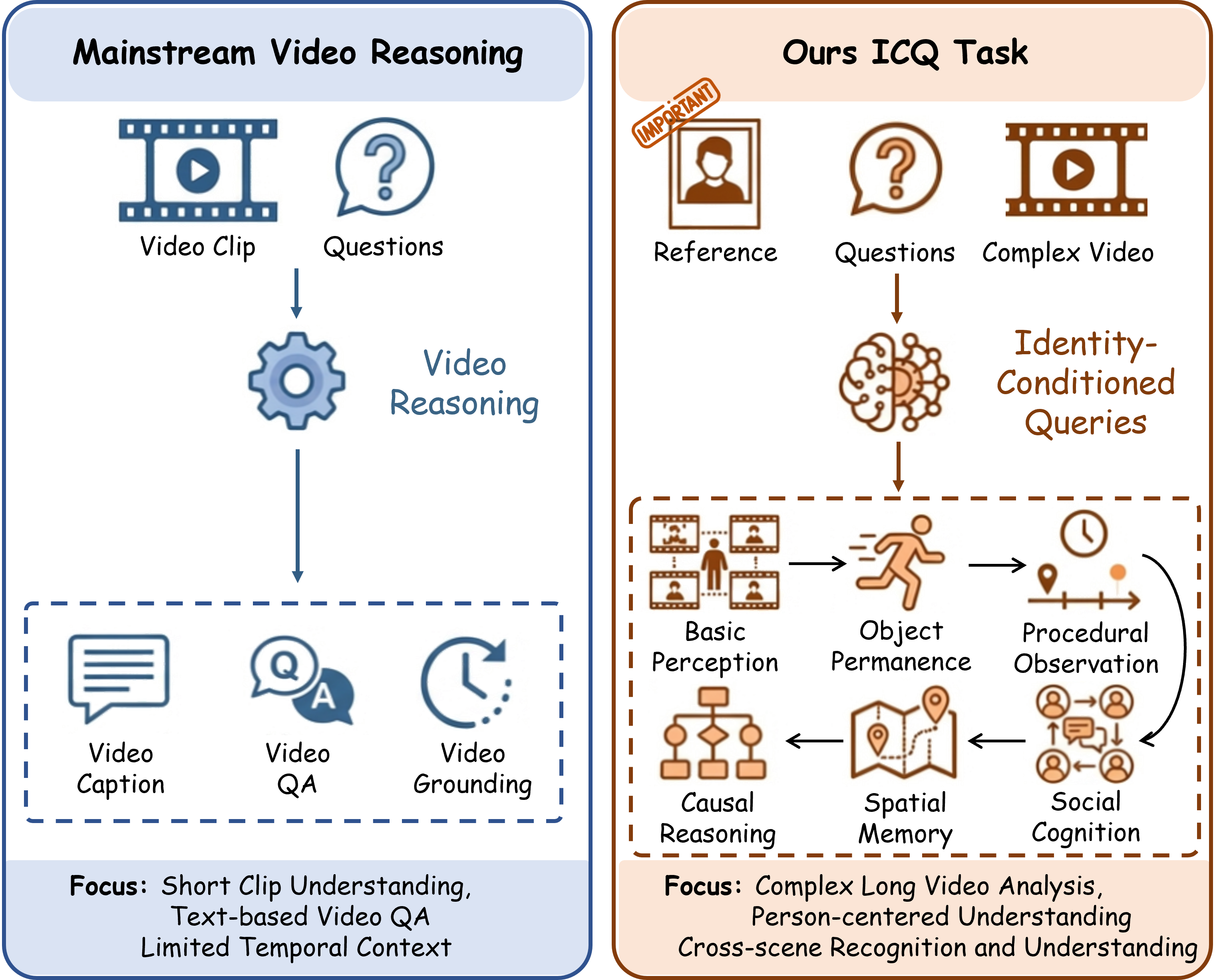}
	\caption{Comparison of the ICQ task with traditional video reasoning. Notably, the ICQ's different levels are anchored to the progression of human cognitive abilities.}
	\Description{Comparison of the ICQ task with traditional video reasoning.}
	\label{fig:abstract}
\end{figure}

\begin{figure}[t]
	\centering
	\includegraphics[width=0.9\linewidth]{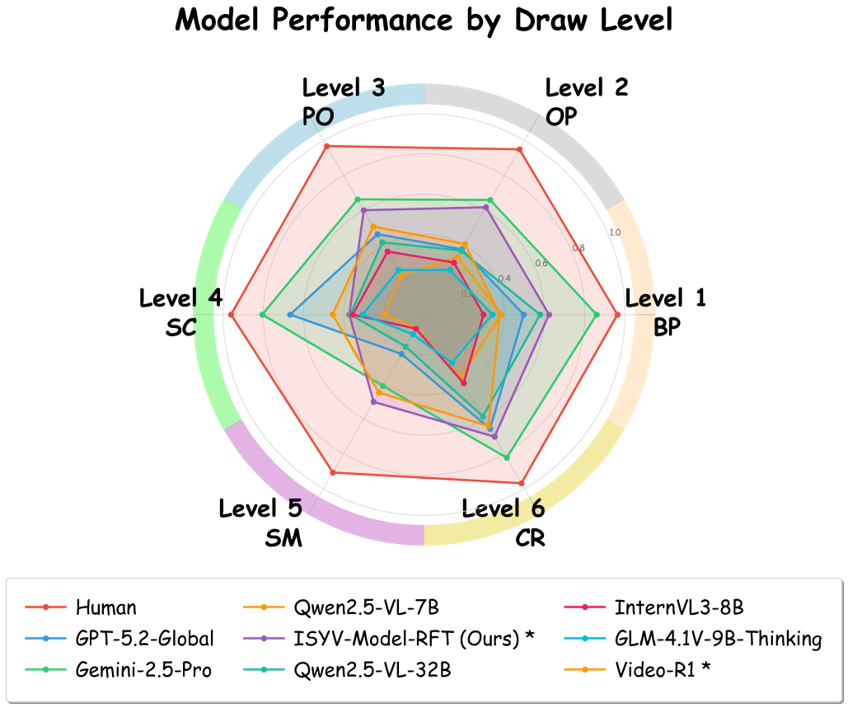}
	\caption{Radar chart of model performance on ISYV-Benchmark across the six levels. The $*$ indicates the model has been trained.}
	\Description{Radar chart of model performance on ISYV-Benchmark across the six levels.}
	\label{fig:radar}
\end{figure}

The comprehensive experiments on ISYV-Bench show that existing mainstream MLLMs perform far from satisfactorily on the ICQ task. As shown in Fig.~\ref{fig:radar}, existing closed-source models still exhibit a substantial gap to human performance on ISYV-Benchmark, while mainstream open-source models perform even worse, with overall accuracy below 40\%. Under the 7B setting, our proposed model increases the average accuracy to 57\%, yielding a significant improvement over the base model and achieving performance that is broadly comparable to closed-source models.
Detailed ablation studies and case analyses reveal key bottlenecks of current MLLMs in cross-domain identity matching and long-horizon tracking, providing important insights for future development in this area.

In summary, our main contributions are as follows:
\begin{itemize}
	\item[$\bullet$] We formulate Identity-Conditioned Queries (ICQ), a practical person-centric task requiring joint reasoning over a video, a reference image of the target person, and a text query.
	\item[$\bullet$] We build a comprehensive data suite for ICQ: ISYV-Bench, a 1,377-sample evaluation benchmark organized into a six-level cognitive hierarchy~\cite{humancogitive}, and ISYV-75K, a 75K-sample training set spanning diverse scenes, identities, and question types.
	\item[$\bullet$] We propose ISYV-Framework, featuring an ICQ Module for reference-image abstraction and an ICQ-oriented training strategy with a novel ESR reward that selects identity-relevant evidence without shot-level annotations. With 7B parameters, it rivals closed-source MLLMs.
	\item[$\bullet$] We comprehensively evaluate mainstream MLLMs on ISYV-Bench, revealing key bottlenecks in cross-domain identity matching and long-horizon tracking, with ablations offering insights for future research.
\end{itemize}

\section{Related Works}
\subsection{Video Reasoning Benchmarks}

Current mainstream video reasoning benchmarks primarily encompass multiple sub-tasks, including Video Question Answering (VQA)~\cite{b-mmvu}, Video Grounding~\cite{b-Charades-STA}, and Video Captioning~\cite{b-vcapsbench}, aimed at comprehensively evaluating models' understanding and analytical capabilities on individual videos~\cite{b-activitynetqa}. Early benchmarks such as TVQA~\cite{b-tvqa} assess models' video comprehension through closed- or open-ended questions on video clips. Subsequently, works like NExT-QA~\cite{b-nextqa} and LongVideoBench~\cite{b-longvideobench} introduce spatial-temporal reasoning challenges, further increasing evaluation complexity. Recently, benchmarks such as MVBench~\cite{b-mvbench} and Video-MME~\cite{b-videomme} have expanded the diversity of task types. However, these benchmarks uniformly adopt a "video-text" dual-modal input paradigm and rarely address videos containing complex shot transitions. Beyond general-purpose video reasoning, domain-specific video analysis has been studied extensively, ranging from video anomaly detection~\cite{suvad,accv-svad,semi-vad3d,vad-survey} to long-term multi-object tracking~\cite{pig-tracking}, where associating and following specific subjects over time is essential; these methods, however, operate over fixed label spaces rather than open-ended, identity-conditioned reasoning. IDA-VLM ~\cite{ida-vlm}, PLVM~\cite{plvm} and others~\cite{iir-vlm,myvlm} attempt to incorporate person images as references to assist models in understanding images containing specific individuals. Nevertheless, this work remains confined to "image-text" paired tasks, without exploring multi-source input scenarios involving "complex videos, cross-shot person references, and textual queries."

\subsection{RL-based Fine-tuning}
With the rapid advancement of MLLMs, numerous studies have significantly enhanced models' reasoning capabilities through RL approaches~\cite{su2025thinking,zhang2025thinking,video-r1,reason}. The recently proposed Group Relative Policy Optimization (GRPO)~\cite{grpo} algorithm stabilizes the training process and improves training efficiency by computing group wise normalized rewards, achieving substantial performance gains in image and video understanding tasks. Furthermore, building upon the principles of GRPO, researchers have proposed various improved training strategies, such as DAPO~\cite{dapo}, SRPO~\cite{srpo}, GFPO~\cite{gfpo} and Multi-GRPO~\cite{multi-grpo}, the latter extending group-relative estimation to multiple reward signals, further advancing the field. In parallel, a complementary line of work reduces the deployment cost of large models through compression, quantization, and distillation~\cite{lbllm,song2025achieving}. However, to date, no study has systematically validated the effectiveness of GRPO and its variants on ICQ or similar tasks.

\begin{table*}[]
	\centering
	\caption{Comparison of our ISYV-Benchmark with mainstream video reasoning datasets. Our benchmark highlights the importance of multimodal input integration and a cognitively inspired hierarchical task design.}
	\resizebox{\linewidth}{!}{
		\begin{tabular}{l|ccccc|ccc}
			\toprule
			Benchmark & \#Video Num & \#QA Pairs & \#Duration(s) & Task Num & Annotation & Complex transitions & Multi-input & Human-Aligned Cognition \\
			\midrule
			TVQA~\cite{b-tvqa} &2,179 &15,253 &11 &3 &M &\xmark &\xmark &\xmark \\
			MVBench~\cite{b-mvbench} &3,641 &4,000 &16 &20 &A &\xmark &\xmark &\xmark \\
			ActivityNet-QA~\cite{b-activitynetqa} &5,800 &58,000 &180 &4 &M &\cmark &\xmark &\xmark \\
			NExT-QA~\cite{b-nextqa} &5,440 &52,044 &44 &2 &M &\xmark &\xmark &\xmark \\
			LongVideoBench~\cite{b-longvideobench} &3,763 &6,678 &473 &17 &M &\cmark &\xmark &\xmark \\
			MMVU~\cite{b-mmvu} &1,529 &3,000 &51 &27 &M &\xmark &\xmark &\xmark \\
			Video-MME~\cite{b-videomme} &900 &2,700 &1,017 &12 &M &\cmark &\xmark &\xmark \\
			MLVU~\cite{b-mlvu} &1,730 &3,102 &930 &9 &M+A &\cmark &\xmark &\xmark \\
			Ego-Exo4D~\cite{b-egoexo4d} &5,035 & / &156 &4 &M &\cmark &\xmark &\xmark \\
			EgoExoLearn~\cite{b-egoexolearn} &747 & / & / &4 &M &\cmark &\xmark &\xmark \\
			CrossVid~\cite{crossvid} &5,331 &9,012 &215 &10 &M+A &\cmark &\xmark &\xmark \\
			\midrule
			ISYV-Benchmark &1,377 &1,377 &50 &17 &M+A &\cmark &\cmark &\cmark \\
			\bottomrule
		\end{tabular}
	}
	\label{tab:benchmark}
\end{table*}

\section{ISYV-75K \& ISYV Benchmark}

\subsection{Overview}
The ISYV-75K dataset comprises 24,150 unique video clips and 74,578 question-answer pairs, with each video accompanied by a corresponding character patch to guide the model in answering questions. The ISYV Benchmark consists of 1,377 high-quality video clips, 1,377 corresponding character patches, and 1,377 question-answer pairs. To ensure data quality, we collected over 100,000 video clips from authentic film and television productions, encompassing a wide range of video durations and varying degrees of visual complexity. During the video selection process, we emphasized shot transition complexity, scene diversity, and character distinctiveness, ensuring that the resulting dataset is sufficiently challenging while remaining suitable for the ICQ task.

\subsection{ISYV-75K Training Dataset}

\subsubsection{ICQ Task Definition.}
As aforementioned, the Identity Conditioned Queries (ICQ) task represents a further extension of current mainstream Video Reasoning in the domain of human-centric cognition. Specifically, the model is required to simultaneously process a long-form video containing complex shot transitions and a reference image of a person, where the latter serves to guide the model in focusing on a specific character when answering questions. The model must comprehend the relationship between the character in the reference image and the video content, thereby accurately answering related questions. Fig.~\ref{fig:abstract} illustrates the distinction between our ICQ task and mainstream Video Reasoning tasks.

\subsubsection{Hierarchical Tasks.}
We position the ICQ task as an extension of Video Reasoning capabilities within human-centric scenarios. Consequently, the ICQ task must encompass as many sub-task types as possible to comprehensively evaluate model performance. Based on this, we align the design philosophy of the ICQ task with the developmental trajectory of human cognitive abilities~\cite{humancogitive}.

\begin{figure}[t]
	\centering
	\includegraphics[width=\linewidth]{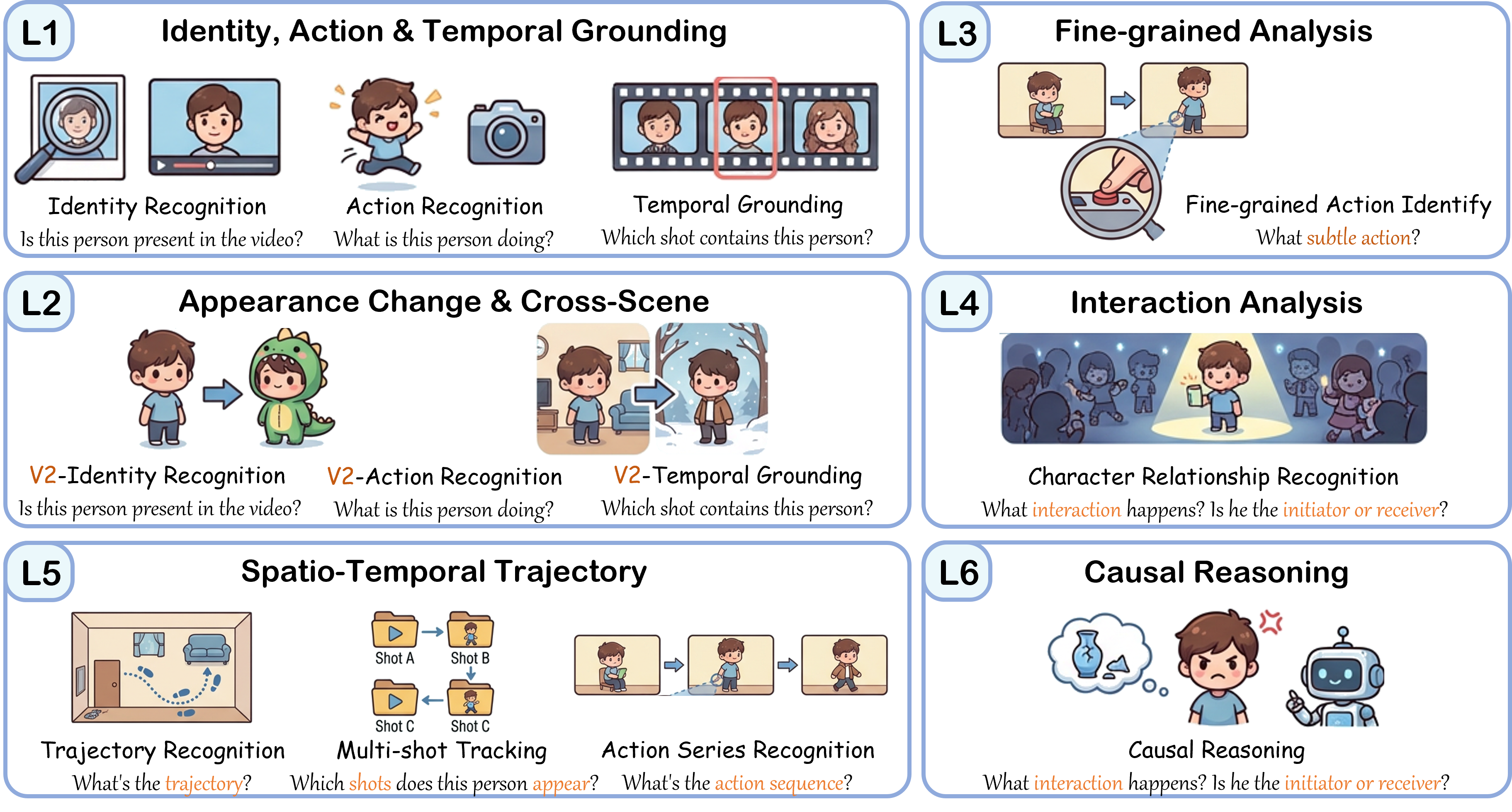}
	\caption{Schematic diagram of the hierarchical level design in ICQ.}
	\Description{Schematic diagram of the hierarchical level design in ICQ.}
	\label{fig:dataset}
\end{figure}

\begin{table}
	\centering
	\caption{Hierarchical levels of the ICQ task and their corresponding human cognitive abilities.}
	\label{tab:levels}
	\resizebox{\columnwidth}{!}{
		\begin{tabular}{ll}
			\toprule
			\textbf{Description} & \textbf{Cognitive Analogy} \\
			\midrule
			L1 - Who, what, and which shot? & Basic Perception~(BP) \\
			L2 - Same person, different outfit or new location & Object Permanence~(OP) \\
			L3 - Subtle details & Procedural Observation~(PO) \\
			L4 - Gaze, dialogue, exchange, and social dynamics & Social Cognition~(SC) \\
			L5 - Movement path within and across shots & Spatial Memory~(SM) \\
			L6 - Inferring `why' and understanding hidden intent & Causal Reasoning~(CR) \\
			\bottomrule
		\end{tabular}
	}
\end{table}

Specifically, we devise six hierarchical task levels with progressively increasing difficulty, where each level comprises several further subdivided question types, and each level corresponds to a specific human cognitive capability. Fig.~\ref{fig:dataset} presents exemplar tasks across different levels. Furthermore, Table.~\ref{tab:levels} elaborates the correspondence between each level and its associated human cognitive ability. Through this design, ICQ enables comprehensive evaluation of model performance and establishes connections between MLLMs and human cognitive science.

\subsubsection{End-to-end automated construction process}
ISYV-Benchmark is constructed through manual curation and verification based on ISYV-75K. We therefore first describe the semi-automatic pipeline for building ISYV-75K. We collected over 100,000 video clips from diverse audiovisual sources, including films, TV series, and anime. During preliminary filtering, clips containing no visible characters were removed. The retained clips were then analyzed in depth by \textit{Gemini-2.5-Pro}~\cite{gemini25}. Specifically, we first applied TransNet V2~\cite{transnetv2} to segment each video with timestamps. \textit{Gemini-2.5-Pro} then generated detailed descriptions for each shot, covering environmental context and ID-tagged character actions. Finally, these per-shot analyses were aggregated, and the model was invoked again to produce comprehensive structured annotations encompassing detailed shot descriptions, character appearance attributes, dialogue and narration, and an overall narrative analysis of the video.

Since character bounding boxes are produced by the VLMs, we employed \textit{Qwen3-VL-32B}~\cite{qwen3vl} to perform multiple rounds of quality checks on each reference image, discarding samples that contain multiple persons, are blurry, or lack any visible character. Drawing on mainstream video reasoning datasets and established human cognitive ability frameworks, we then designed the aforementioned task categories along with corresponding QA templates~\cite{b-videotree,b-drvideo}. \textit{Qwen3-Max}~\cite{qwen3} was subsequently used to generate question–answer pairs conditioned on the annotations and QA templates, while descriptions of the reference images were produced by \textit{Qwen3-VL-32B}. All resulting data underwent a plausibility check via \textit{Qwen3-32B}~\cite{qwen3}, after which \textit{Qwen3-Max} was invoked again to generate CoT rationales and to perform leakage detection.

To generate outfit-changed and cross-shot character images, we adopted two complementary strategies. The first leverages the \textit{Qwen-Image-Edit}~\cite{qwenimage} to selectively modify visual attributes of a character (e.g., clothing, accessories, hair color, and pose) while preserving the background, thereby producing diverse appearance variants of the same identity. The second exploits established face ReID techniques~\cite{ren2023pbidr,an_2022_pfc_cvpr} to retrieve additional appearances of the same character from the same source material, selecting reference images that exhibit maximal differences in pose and background. To ensure identity consistency after editing or replacement, we used \textit{Qwen3-VL-32B}~\cite{qwen3vl} and \textit{Qwen3-32B}~\cite{qwen3} to independently describe, compare, and verify each image pair, retaining only samples that passed both visual and textual cross-checks.

\begin{figure}[t]
	\centering
	\includegraphics[width=\linewidth]{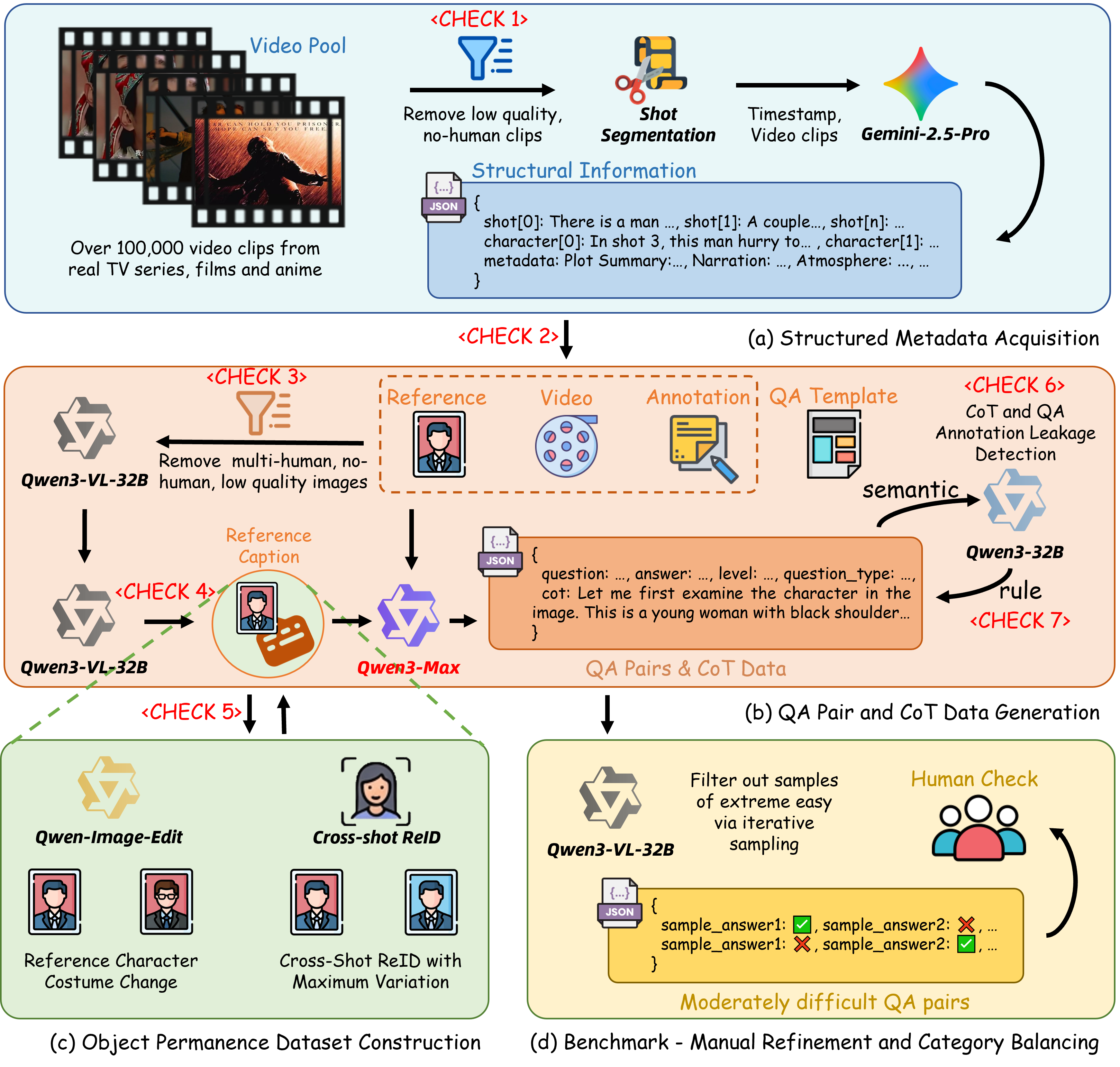}
	\caption{A schematic illustration of our pipeline for constructing ISYV-Benchmark and ISYV-75K. (a), (b), and (c) demonstrate the complex yet effective automated pipeline for constructing the foundational training set. We integrate multi-round checking mechanisms <check X> into the automated data construction pipeline. Detailed information for each check is provided in the appendix.}
	\Description{A schematic illustration of the pipeline for constructing ISYV-Benchmark and ISYV-75K with multi-round checking mechanisms.}
	\label{fig:annotation}
\end{figure}

Fig.~\ref{fig:annotation} (a)-(c) illustrate the construction pipeline of ISYV-75K in detail. 
To ensure the high quality of data obtained through the automated annotation pipeline, we incorporated multiple quality checks after each critical step. The specific locations of these checks are marked as \texttt{\seqsplit{<check X>}}. We will present a more comprehensive description of the pipeline, along with the settings and prompts for the annotation process, \textbf{in the appendix}.

\subsection{ISYV Benchmark}

\subsubsection{Data Selection and Manual Verification.}
To construct a high-quality benchmark, we employ \textit{Qwen3-VL-32B}~\cite{qwen3vl} to perform multiple rounds of inference on the candidate dataset. After filtering out samples that are excessively simple, we engage a professional annotation team to manually validate the category-balanced filtered data. This process yields a benchmark comprising 1,377 high-quality samples, each consisting of a video, character reference images, and corresponding QA pairs. Notably, since ISYV-75K has undergone multiple rounds of automated quality inspection during its construction, the processing pipeline for ISYV-Benchmark is considerably streamlined, which further corroborates our dual quality assurance for both the training and evaluation sets.

Notably, we enforced strict quality control throughout the manual annotation process. Specifically, each sample was cross-validated, error-analyzed, and revised by at least three annotators, followed by a second-round spot check after the completion of the full review process to ensure the accuracy and consistency of the annotations.

\subsubsection{Comparison with Existing Benchmarks.}
We conduct a comprehensive comparison between mainstream Video Reasoning benchmarks and our proposed ISYV Benchmark, with results summarized in Table.~\ref{tab:benchmark}. In contrast to the prevalent "video-text" input paradigm, ISYV Benchmark innovatively introduces the ICQ mechanism, which requires models to simultaneously process video content and character reference images, thereby enabling person-centric fine-grained video reasoning. ISYV significantly extends the scope of cognitive hierarchy design, question type diversity, and application scenario complexity.

Specifically, while existing benchmarks primarily focus on general video reasoning capabilities, ISYV concentrates on person-centric long-form video reasoning, with particular emphasis on cross-shot and cross-scene character tracking and behavior analysis. Our proposed six-level cognitive hierarchy—ranging from basic character recognition to high-level intent inference—aligns with the human cognitive development process, filling the gap in cognitive science-oriented evaluation within current benchmarks. Furthermore, the challenging samples in ISYV Benchmark, including complex shot transitions, character costume changes, and cross-scene tracking, pose more rigorous requirements for models' spatiotemporal reasoning capabilities and fine-grained visual understanding abilities.


\section{ISYV Framework}

In this section, we present a detailed description of the ISYV Framework, which we have developed specifically for ICQ task. We begin by introducing the ICQ Module, a component that leverages shorter learnable token sequences to capture key representations of reference images, thereby enhancing the model's capability for multimodal input alignment. Subsequently, we elaborate on our training strategy and the specifically designed reward function.

\subsection{ICQ Module}

ICQ extends conventional video reasoning by introducing an additional person reference image, making the effective processing of the reference image and its alignment with the video content a central challenge. Although mainstream multimodal large language models support joint video–image inputs, directly reusing the original visual encoder and simply concatenating video tokens with image tokens leads to two notable issues. First, the model often fails to correctly capture the semantic relationship between the reference image and the video, frequently misinterpreting the reference image as the final frame of the video sequence, thereby conflating the logical relations among inputs. Second, the token sequence produced by the image encoder is typically excessively long, which not only introduces irrelevant noise (e.g., background textures and illumination variations) but also incurs unnecessary computational overhead.

\begin{figure}[t]
	\centering
	\includegraphics[width=\linewidth]{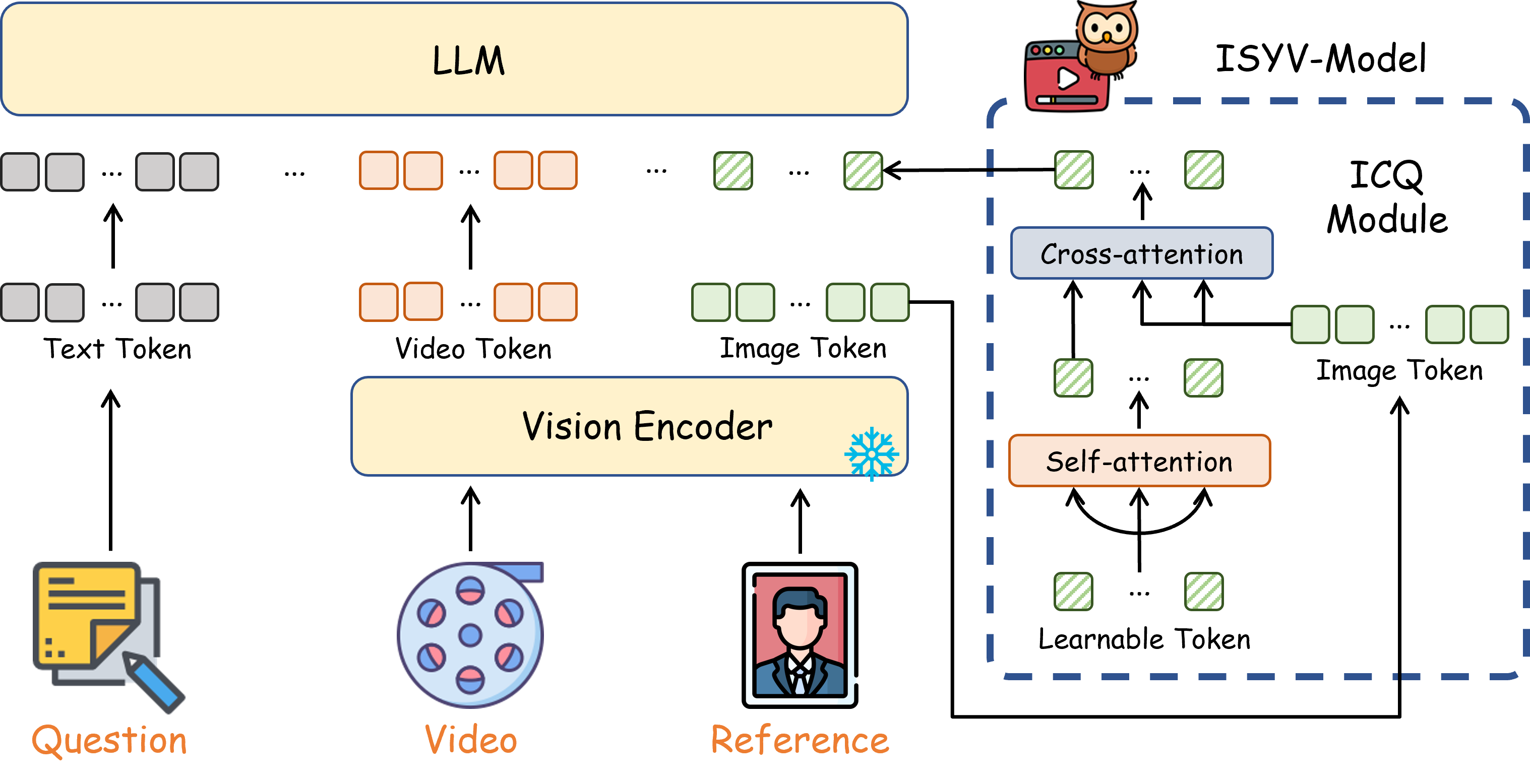}
	\caption{A schematic illustration of our ICQ Module architecture and its integration with MLLM. 
		}
	\Description{A schematic illustration of the ICQ Module architecture and its integration with MLLM.}
	\label{fig:model}
\end{figure}

To address these limitations, we propose the ICQ Module, a simple yet effective architecture. Its core mechanism lies in compressing the reference image into a compact representation through learnable tokens and facilitating interaction with image features via attention mechanisms, thereby extracting person-specific features that are maximally relevant to the video content~\cite{blip2,qformer-vision}.

Concretely, the workflow is as follows. We first initialize a short sequence of learnable tokens and apply self-attention to model their internal dependencies:
\begin{equation}
	\mathcal{T}_{icq}^{(1)} = \mathbf{SelfAttn}(\mathcal{T}_{icq}^{(0)}, \mathcal{T}_{icq}^{(0)}, \mathcal{T}_{icq}^{(0)}), \mathcal{T}_{icq}^{(0)} \in \mathbf{R}^{N_q \times d},
\end{equation}
where $\mathcal{T}_{icq}^{(0)}$ represents the initialized learnable tokens sequence, $\mathbf{SelfAttn}(\cdot)$ denotes the self-attention mechanism, $N_q$ is the number of learnable tokens, and $d$ is the hidden dimension. 

Next, we use these learnable tokens as Queries and the image token sequence output by the original image encoder as Keys and Values, performing cross-attention to obtain an initial representation of the reference image:
\begin{equation}
	\mathcal{T}_{icq}^{(2)} = \mathbf{CrossAttn}(\mathcal{T}_{icq}^{(1)}, \mathcal{T}_{image}, \mathcal{T}_{image}), \mathcal{T}_{image} \in \mathbf{R}^{N_i \times d},
\end{equation}
where $\mathbf{CrossAttn}(\cdot)$ denotes the cross-attention mechanism, $\mathcal{T}_{image}$ represents the original image token sequence, $N_i$ is the number of image tokens from original vision encoder.

Overall, ICQ Module compresses the reference image into a compact representation using learnable tokens, alleviating the semantic confusion and computational inefficiency caused by naive video–image token concatenation. A schematic illustration of our proposed ICQ-Module and the overall ISYV-Model architecture is provided in Fig.~\ref{fig:model}.

\subsection{Two Step Training Strategy}

Regarding the training strategy, we follow mainstream post-training methodologies and adopt a two-stage pipeline~\cite{zhang2025tinyllava, wang2025time, tian2025ego}. We first perform supervised fine-tuning as a cold-start phase, enabling the model to acquire foundational knowledge and learn the target output format. Subsequently, we conduct reinforcement fine-tuning using the GRPO algorithm to further enhance model performance.

To construct data suitable for different training stages, we first perform quality filtering on the chain-of-thought samples in ISYV-75K. Specifically, we deploy \textit{Qwen3-32B}~\cite{qwen3} locally to score each CoT sample on a scale from 0 to 10 based on reasoning quality and label leakage risk, where a score of 10 indicates no label leakage with sound and complete reasoning, while a score of 0 indicates evident label leakage, such as explicitly referencing the original video annotations during the reasoning process.

Based on these scores, we partition the data into SFT and RFT subsets with an intra-class ratio of 6:4. Since the SFT stage imposes explicit supervision on the reasoning component, leaked information in CoT may directly contaminate the supervision signal. Therefore, we avoid allocating low-scoring samples to the SFT.

\subsection{Reward Function}

Mainstream video reasoning methods typically employ answer accuracy reward $r_{acc}$ and format accuracy reward $r_{fmt}$ as signals for reinforcement learning~\cite{park2025deepvideo,fang2025viss}. 
However, such approaches are far less effective on long videos with complex scene transitions, and especially on ICQ: the model struggles to locate question-relevant shots and instead processes the whole video, introducing substantial noise. Yet annotating effective shots is labor-intensive and highly subjective.

To address this, beyond the above signals we design an Effective Shot Re-Reasoning (ESR) reward on top of GRPO and a caption reward for the reference image. The ESR reward triggers secondary re-reasoning on the initial sampling results, teaching the model which shots are effective without ground-truth shot annotations.

Specifically, our RFT process employs the following four reward functions:

\textbf{Format Reward.}
To facilitate more intuitive evaluation and training for the ICQ task, we decompose it into four components, corresponding to the four required output fields.
Specifically, \texttt{\seqsplit{<caption>...</caption>}} summarizes the target task based on the reference image, \texttt{\seqsplit{<candidate>...</candidate>}} identifies the relevant shots for the query, and \texttt{\seqsplit{<think>...</think>}} and \texttt{\seqsplit{<answer>...</answer>}} denote the reasoning process and final prediction, respectively. This decomposition enables the model to learn reference alignment, evidence localization, reasoning, and answer generation, thereby improving task performance while mitigating answer hacking.
The format reward is defined as:
\begin{equation}
	r_{fmt} = \begin{cases} 
		1.0, & \text{all tags and in correct order} \\ 
		0.9, & \text{all tags but in wrong order} \\ 
		\sum_{t} w_t \cdot \mathbf{1}_t, & \text{otherwise} 
	\end{cases},
\end{equation}
where $w_{cap}$ = 0.2, $w_{cand}$ = 0.15, $w_{think}$ = 0.15, $w_{ans}$ = 0.5, and $\mathbf{1}_t$ denotes the indicator function for the presence of tag $t$.

\textbf{Accuracy Reward.}
We employ a binary score for all task types, evaluating the consistency between the answer $A_1$ and the ground-truth answer $A_{gt}$:
\begin{equation}
	r_{acc} = \begin{cases} 
		1.0, & \text{if}~A_1 = A_{gt} \\ 
		0.0, & \text{otherwise} \\ 
	\end{cases}.
\end{equation}

\textbf{Caption Semantic Reward.}
We employ an LLM $\mathcal{M}_{judge}$ as a judge to evaluate the semantic consistency between the model-generated caption $C_{pred}$ and the ground-truth reference image description $C_{gt}$, with particular emphasis on identifying obvious conflicts. The judge model outputs a score ranging from 0 to 10, which is mapped to a reward through a step function:
\begin{equation}
	r_{cap} = \omega \times \lfloor s/2 \rfloor, s = \mathcal{M}_{judge}(C_{pred}, C_{gt})\in [0,10],
\end{equation}
where $\omega$ is the Hyperparameter.
This reward facilitates better understanding of the reference image content related to the question, preventing the model from exploiting shortcuts by analyzing only the video and text while ignoring the reference image. Furthermore, this approach implicitly supervises the think process to ensure more reasonable reasoning.

\textbf{Effective Shot Re-Reasoning Reward.}
During the GRPO rollout phase, the policy model generates an output containing the effective shot sequence
$S = \{s_1,s_2,...,s_k\}$, chain-of-thought $T$, and final answer $A_1$. Let $N_{ref}$ denote the total number of shots containing the reference person in the video, and $k=|S|$ denote the number of shots output by the model. If the initial answer is correct and the shot sequence can be successfully parsed, we clip and concatenate the shots mentioned by the model according to timestamp annotations, reduce the overall sampling frame rate, and perform re-reasoning under the same settings to obtain the re-reasoning answer $A_2$.  The re-reasoning reward is defined as:
\begin{equation}
	r_{esr} = \mathbf{1}[A_1 = A_{gt}] \cdot \mathbf{1}[\mathbf{Parse}(S)] \cdot \mathbf{1}[A_2 = A_{gt}] \cdot (0.5 + r_{extra}),
\end{equation}
where the extra reward $r_{extra}$ decreases linearly with the number of output shots:
\begin{equation}
	r_{extra} = \begin{cases} 
		0.5, & \text{if}~k=1 \\ 
		0, & \text{if}~k\geq N_{ref}\\ 
		0.5 \times \frac{N_{ref} - k}{N_{ref} - 1}, & \text{otherwise} \\ 
	\end{cases}.
\end{equation}
For certain types of questions, there exist cases where $N_{ref} = \mathbf{None}$. In such cases, the ESR Reward degenerates to a binary score consistent with the accuracy reward.
This mechanism enables the model to learn how to identify effective shots and achieve localization and acquisition of key information without ground-truth effective shot annotations.

In summary, the overall reward function is:
\begin{equation}
	r = \theta_{acc} \cdot r_{acc} + \theta_{fmt} \cdot r_{fmt} + \theta_{cap} \cdot r_{cap} + \theta_{esr} \cdot r_{esr},
\end{equation}
where $\theta_{acc}, \theta_{fmt}, \theta_{cap}$ and $\theta_{esr}$ are hyperparameters. We will discuss these \textbf{in the appendix}.

It is worth noting that no effective shot annotations are included during the data construction phase. Therefore, in the SFT stage, we train the model to output all shots containing the reference person, and this output will be gradually optimized to effective shots that contribute to answering the question during the RFT stage.

\begin{table*}[]
	\centering
	\caption{Evaluation results of mainstream closed-source models, open-source models, and related works on ISYV-Benchmark. $O. Avg$ represents the accuracy of correct answers only, while $ICQ~Q.Avg$ represents the accuracy under ICQ's unique requirements where both image descriptions and correct answers are verified. The blue values reflect the gap between the two. $\dag$ denotes full adherence to the original training setup, $\ddag$ denotes the use of our training setup with the model architecture unchanged. $\star$ The evaluation was conducted blindly by three annotators.}
	\resizebox{\linewidth}{!}{
		\begin{tabular}{c|cc|cc|cccccc}
			\toprule
			\textbf{Models} & \textbf{Query Format} & \textbf{Think} & \textbf{$O.Avg$} & \textbf{$ICQ~Q.Avg$} & \textbf{Level 1 - BP} & \textbf{Level 2 - OP} & \textbf{Level 3 - PO} & \textbf{Level 4 - SC} & \textbf{Level 5 - SM} & \textbf{Level 6 - CR} \\
			\midrule
			Human$\star$ &Image & / &95.13\% & / &96.30\% &95.02\% &96.88\% &96.11\% &90.81\% &96.97\% \\
			\midrule
			GPT-5.2-Global~\cite{gpt5} &Image &Yes &45.53\% &45.53\% \textcolor{blue}{(-0.00\%)} &49.63\% &37.55\% &46.25\% &66.67\% &22.61\% &65.66\% \\
			Gemini-2.5-Pro~\cite{gemini25} &Image &Yes &\underline{\textbf{67.10\%}} &\underline{\textbf{67.10\%}} \textcolor{blue}{(-0.00\%)} &\underline{\textbf{85.93\%}} &\underline{\textbf{65.90\%}} &\underline{\textbf{66.25\%}} &\underline{\textbf{80.56\%}} &\underline{\textbf{40.99\%}} &\underline{\textbf{82.32\%}} \\
			Qwen3-VL-235B-A22B~\cite{qwen3vl} &Image &No &39.80\% &39.80\% \textcolor{blue}{(-0.00\%)} &70.37\% &45.59\% &37.81\% &27.78\% &21.91\% &51.01\% \\
			\midrule
			Qwen2.5-VL-7B~\cite{qwen25vl} &Text &No &22.44\% &/ &42.96\% &24.90\% &21.88\% &19.44\% &8.48\% &28.79\% \\
			Qwen2.5-VL-32B~\cite{qwen25vl} &Text &No &\underline{\textbf{38.56\%}} &/ &\underline{\textbf{55.56\%}} &\underline{\textbf{38.70\%}} &\underline{\textbf{39.69\%}} &\underline{\textbf{33.89\%}} &19.43\% &\underline{\textbf{56.57\%}} \\
			Qwen3-VL-8B~\cite{qwen3vl} &Text &No &28.61\% &/ &55.56\% &33.33\% &20.00\% &6.11\% &\underline{\textbf{28.62\%}} &38.38\% \\
			\midrule
			Qwen2.5-VL-7B~\cite{qwen25vl} &Image &No &24.76\% &6.75\% \textcolor{blue}{(-18.01\%)} &38.52\% &33.33\% &22.50\% &20.00\% &7.77\% &36.36\% \\
			Qwen2.5-VL-32B~\cite{qwen25vl} &Image &No &\underline{\textbf{39.29\%}} &\underline{\textbf{33.99\%}} \textcolor{blue}{(-5.30\%)} &\underline{\textbf{57.78\%}} &\underline{\textbf{36.78\%}} &41.56\% &36.67\% &18.37\% &\underline{\textbf{58.59\%}} \\
			Qwen3-VL-8B~\cite{qwen3vl} &Image &No &31.45\% &23.09\% \textcolor{blue}{(-8.36\%)} &44.44\% &32.18\% &28.75\% &17.22\% &\underline{\textbf{23.67\%}} &50.00\% \\
			GLM-4.1V-9B-Thinking~\cite{glm41v} &Image &Yes &24.47\% &22.80\% \textcolor{blue}{(-1.67\%)} &34.07\% &25.67\% &25.62\% &30.56\% &11.31\% &27.78\% \\
			Mimo-VL-7B-RL~\cite{mimovl} &Image &Yes &18.74\% &10.09\% \textcolor{blue}{(-8.65\%)} &20.00\% &17.24\% &22.81\% &20.00\% &10.60\% &23.74\% \\
			InternVL3-8B~\cite{internvl3} &Image &No &28.98\% &28.90\% \textcolor{blue}{(-0.08\%)} &29.63\% &29.89\% &36.25\% &35.56\% &8.13\% &39.39\% \\
			VideoLLaMA3-7B~\cite{videollama3} &Image &No &37.25\% &3.34\% \textcolor{blue}{(-33.92\%)} &34.07\% &35.25\% &\underline{\textbf{47.81\%}} &\underline{\textbf{40.00\%}} &16.61\% &52.02\% \\
			\midrule
			IDA-VLM~\cite{ida-vlm} $^{\dag}$ &Image &No &25.05\% &24.84\% \textcolor{blue}{(-0.21\%)} &31.11\% &31.03\% &15.94\% &5.56\% &36.75\% &28.79\%\\
			PLVM~\cite{plvm} $^{\dag}$ &Image &No &10.97\% &5.16\% \textcolor{blue}{(-5.81\%)} &5.93\% &6.90\% &14.06\% &30.00\% &5.65\% &5.05\%\\
			Video-R1~\cite{video-r1} $^{\dag}$ &Image &Yes &47.57\% &47.57\% \textcolor{blue}{(-0.00\%)} &37.78\% &40.61\% &50.62\% &\underline{\textbf{45.56\%}} &44.88\% &64.14\% \\
			Qwen2.5-VL-7B-SFT~\cite{qwen25vl} $^{\ddag}$ &Image &No &28.03\% &28.03\% \textcolor{blue}{(-0.00\%)}  &37.04\% &36.40\% &19.38\% &8.33\% &37.46\% &29.29\% \\
			Qwen2.5-VL-7B-RFT~\cite{qwen25vl} $^{\ddag}$ &Image &Yes &54.22\% &54.22\% \textcolor{blue}{(-0.00\%)} &59.25\% &51.72\% &53.12\% &34.44\% &49.82\% &69.19\% \\
			\underline{ISYV-Model-SFT(Ours)} &Image &No &33.55\% &33.40\% \textcolor{blue}{(-0.15\%)} &49.63\% &46.74\% &20.31\% &8.33\% &43.82\% &34.85\% \\
			\underline{ISYV-Model-RFT(Ours)} &Image &Yes &\underline{\textbf{57.01\%}} &\underline{\textbf{57.01\%}} \textcolor{blue}{(-0.00\%)} &\underline{\textbf{62.22\%}} &\underline{\textbf{61.69\%}} &\underline{\textbf{60.00\%}} &37.22\% &\underline{\textbf{50.18\%}} &\underline{\textbf{70.20\%}} \\
			\bottomrule
		\end{tabular}
	}
	\label{tab:main}
\end{table*}

\section{Experiments and Analysis}

This section presents a systematic experimental evaluation and analysis of current mainstream open-source and closed-source MLLMs, as well as our specifically designed ISYV-Model for the ICQ task, on the ISYV-Benchmark. We first introduce the experimental setup, then analyze the performance of each model on the benchmark, and finally conduct ablation studies and case studies to investigate the impact of key factors in depth.

\subsection{Experimental Setup}
We evaluated multiple mainstream MLLMs on the ISYV-Benchmark, encompassing closed-source models (e.g., GPT-5.2-global~\cite{gpt5}, Gemini-2.5-Pro~\cite{gemini25}), open-source models (e.g., Qwen2.5-VL~\cite{qwen25vl}, Qwen3-VL~\cite{qwen3vl}, InternVL3~\cite{internvl3}), and our proposed ISYV-Model, with parameter scales ranging from 7B to hundreds of billions. For the video component in multi-source inputs, we uniformly distributed the total number of sampled frames across the entire video duration and standardized the frame sampling to 32 frames for all evaluations. Regarding the organization of multi-source inputs, we constructed joint video-image-text inputs according to the specifications provided by the transformers library or the official documentation of each model, ensuring that all modality information was fed into the model within a single conversational turn. 

We uniformly required all models to produce outputs following the standard format of the ICQ task: first generating a description of the reference image, then providing a candidate sequence of valid shots, and finally outputting the thinking process and the final answer separately. Models supporting the thinking mode were required to output the think component, while those without this capability only needed to output the answer component. 
Inference for open-source models was conducted following their official implementations, while closed-source models were accessed through official APIs. Accuracy was adopted as the primary evaluation metric, and additional implementation details, evaluation settings, \textbf{and training configurations are provided in the appendix.} For IDA-VLM, PLVM, and Video-R1, we retained their original architectures and configurations. For Qwen2.5-VL-7B-SFT/RFT, we concatenated the video and image tokens before feeding them into the LLM, and enabled all reward functions during the RFT stage.

\subsection{Evaluation on the ISYV-Benchmark}

As shown in Table.~\ref{tab:main}, we present the evaluation results of several MLLMs on ISYV-Benchmark, including accuracy across different levels, overall average accuracy, and $ICQ~Q.Avg$. Notably, $ICQ~Q.Avg$ in the table represents the accuracy of responses that are both correct and satisfy the ICQ task requirements (i.e., containing at least an analysis of the reference image and the answer). This metric is necessary because some models tend to completely ignore the input reference image and attempt to hack the answer using only the video and question. We employ rule-based methods or LLM outputs to parse responses and determine whether they contain descriptions of the reference image; \textbf{detailed settings are provided in the appendix}.

Based on these results, we derive several key observations:

\textbf{The ICQ task poses significant challenges for existing MLLMs.}
Current MLLMs exhibit a substantial performance gap compared to humans on ICQ and similar complex tasks. Even the best-performing proprietary model, \textit{Gemini-2.5-Pro}~\cite{gemini25}, falls considerably short of human performance. Humans demonstrate clear advantages across all difficulty levels, particularly in Level 2 tasks involving disguised character analysis.
This reveals a critical limitation of existing MLLMs in jointly analyzing and comprehending complex multi-source inputs.

\textbf{Closed-source MLLMs significantly outperform open-source models and better adapt to the ICQ task.}
All closed-source MLLMs achieve higher overall average accuracy than open-source MLLMs, with their advantages being particularly pronounced on several critical tasks. Furthermore, closed-source models are less inclined to attempt hacking answers and instead tend to correlatively understand the video and reference image. In contrast, \textit{VideoLLaMA3-7B}~\cite{videollama3} achieves a reasonable $O.Avg$ but exhibits catastrophic performance on $ICQ~Q.Avg$. Our analysis of the model outputs reveals that this model fails to comprehend the ICQ task entirely and tends to provide brief responses that appear "random" or attempt to hack an answer. 
Fig.~\ref{fig:casestudy} presents a concrete example of answer hacking from \textit{Qwen2.5-VL-32B}~\cite{qwen25vl}, additional analyses are provided \textbf{in the appendix}.

\textbf{Compared to multi-turn input and text-substituted image approaches, the ICQ task is irreplaceable.}
As previously discussed, the ICQ task cannot be effectively addressed through intuitive alternative methods, such as employing multi-turn dialogue or performing reasoning after substituting images with textual descriptions. Multiple models exhibit significant performance degradation when using text-substituted image inputs. Therefore, the ICQ task represents a necessary attempt to extend the boundaries of the video reasoning domain.

\subsection{Ablation Studies}

\begin{table}[t]
	\centering
	\caption{Ablation studies examining the impact of different training strategies and reward function combinations on model performance. Columns are cumulative: each RFT column adds the listed reward to the previous one.}
	\resizebox{\columnwidth}{!}{
	\begin{tabular}{lccccc}
		\toprule
		\textbf{Strategy} & SFT & SFT-$GT_{cand}$ & RFT $r_{fmt},r_{acc}$ & $+r_{cap}$ & $+r_{esr}$\\ 
		\midrule
		\textbf{$O.Avg$} &33.75\% &35.62\% &52.06\% &55.44\% &\underline{\textbf{57.01\%}} \\
		\bottomrule
	\end{tabular}
	}
	\label{tab:reward}
\end{table}

We first conduct ablation studies on the training strategies and the designed reward functions, with results presented in Table.~\ref{tab:reward}.
The results indicate that omitting the learning of \texttt{\seqsplit{<candidate>...</candidate>}} during the SFT stage yields better immediate performance but constrains the model's ability to learn effective shots in the subsequent RFT stage.
During the RFT stage,  in addition to the standard $r_{fmt}$ and $r_{acc}$, both $r_{cap}$ and $r_{esr}$ further enhance model performance. The $r_{cap}$ facilitates better association between reference image descriptions and the reasoning process, while $r_{esr}$ enables the model to more effectively identify key shot sequences in the video that are relevant to solving the problem.
We require the model to output the \texttt{\seqsplit{<candidate>...</candidate>}} field during the SFT stage only when $r_{esr}$ is enabled.

Furthermore, we conducted ablation studies on the length of the learnable token sequence in the ICQ Module and the number of input frames, with results shown in Table.~\ref{tab:num}. The results indicate that 32 is the optimal token length: too few tokens fail to adequately represent the reference person image, while too many may introduce noise. Regarding the number of input frames, since 32-frame sampling was used during training, increasing the number of frames did not improve performance, whereas using fewer frames led to performance degradation due to insufficient information.

\begin{table}[]
	\centering
	\caption{Ablation studies investigating the effect of varying the length of learnable token sequences in the ICQ Module.}
	\begin{tabular}{cccc}
		\toprule
		\textbf{Token Nums} & 16 Token & 32 Token &64 Token\\
		\midrule
		\textbf{$O.Avg$} &46.93\% &\underline{\textbf{57.01\%}} &53.31\% \\
		\midrule
		\textbf{Frame Nums} & 16 Frames & 32 Frames &64 Frames\\
		\midrule
		\textbf{$O.Avg$} &34.50\% &\underline{\textbf{57.01\%}} &56.56\% \\
		\bottomrule
	\end{tabular}
	\label{tab:num}
\end{table}

\subsection{Case Study}

\begin{figure}[t]
	\centering
	\includegraphics[width=\linewidth]{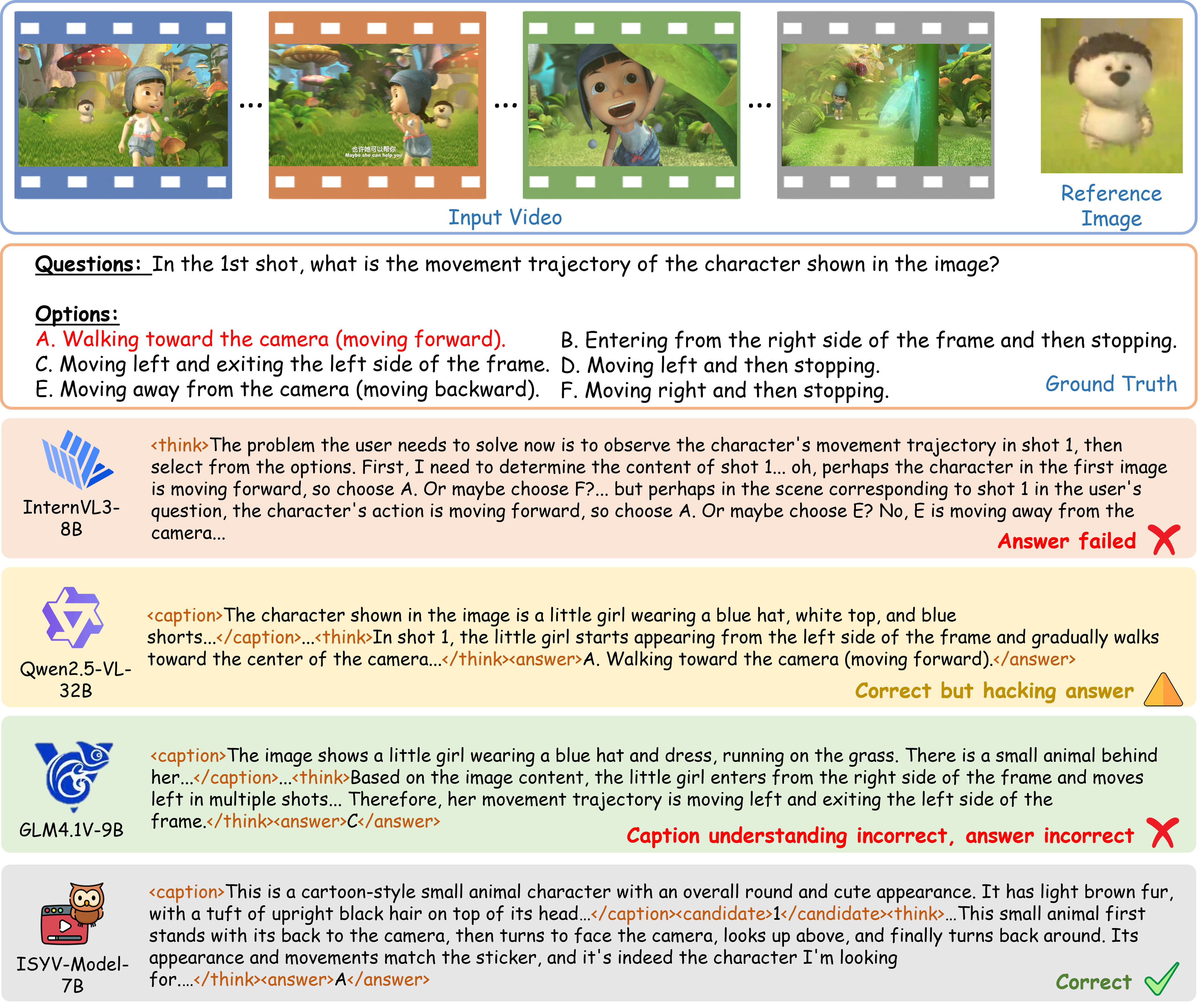}
	\caption{The case study of different models on a sample shows that our model completed the task very well, while other models all exhibited some errors to varying degrees.}
	\Description{The case study of different models on a sample from ISYV-Benchmark.}
	\label{fig:casestudy}
\end{figure}

Fig.\ref{fig:casestudy} shows the responses of different models to a level-5 sample from ISYV-Benchmark. As shown, all baseline models exhibit varying issues. \textit{GLM4.1V-9B}\cite{glm41v} and \textit{Qwen2.5-VL-32B}\cite{qwen25vl} misinterpreted the reference image, \textit{Qwen3-VL-8B}\cite{qwen3vl} incorrectly claimed that the referenced person was absent, and \textit{InternVL3-8B}~\cite{internvl3} failed to follow the required format. Although \textit{Qwen2.5-VL-32B} selected the correct option, this may be considered a form of hacking behavior because its image description was inaccurate. In contrast, our model accurately described the reference person and answered correctly.

\section{Conclusion}
To address the diverse input scenarios in real-world video reasoning, we introduce ICQ, the person-centric multi-input video reasoning task. To be specific, we develop a comprehensive solution encompassing: a manually annotated high-quality evaluation benchmark ISYV-Benchmark (spanning 6 difficulty levels and 17 subtasks), an automatically constructed large-scale training set ISYV-75K, and a dedicated model architecture with corresponding training methodology. Extensive evaluation reveals that the ICQ task poses significant challenges for current MLLMs, with even the best-performing proprietary models exhibiting substantial gaps compared to human-level performance. Through comprehensive experiments and analysis, we uncover fundamental limitations of existing models. We anticipate that the ICQ task will serve as a valuable resource for advancing the boundaries of video reasoning and promoting the development of MLLMs. 
In the future, we will further explore and expand video reasoning tasks to improve MLLMs' capabilities in more complex and realistic scenarios.

\begin{acks}
This work was supported in part by the Key-Area Research and Development Program of Guangdong Province (2024B0101040008).
\end{acks}

\bibliographystyle{ACM-Reference-Format}
\bibliography{icq}

\end{document}